\PassOptionsToPackage{square,comma,numbers,sort&compress,super}{natbib}
\documentclass{article}
\usepackage{amsthm}
\usepackage[table,xcdraw]{xcolor}
\usepackage{algorithm}
\usepackage{algorithmic}
\usepackage[utf8]{inputenc} % allow utf-8 input
\usepackage[T1]{fontenc}    % use 8-bit T1 fonts
\usepackage{url}            % simple URL typesetting
\usepackage{booktabs}       % professional-quality tables
\usepackage{amsfonts}       % blackboard math symbols
\usepackage{nicefrac}       % compact symbols for 1/2, etc.
\usepackage{microtype}      % microtypography
\usepackage[fleqn]{amsmath}
\usepackage{graphicx}
\usepackage{enumitem}
\usepackage{tabularx}
\usepackage{tablefootnote}
\usepackage{authblk}
\usepackage{mathtools}
\usepackage{float}
\usepackage{placeins}
\usepackage[most]{tcolorbox}

\definecolor{pnasbluetext}{RGB}{0,101,165}
\definecolor{pnasblueback}{RGB}{205,217,235}

\usepackage[preprint]{neurips_2024}

\usepackage{hyperref}
\hypersetup{colorlinks,allcolors=black}

\newtheorem{theorem}{Theorem}%  meant for continuous numbers
\newtheorem{conjecture}[theorem]{Conjecture}
\newtheorem{remark}{Remark}%
\newtheorem{definition}{Definition}%

\title{Learning-Based TSP-Solvers Tend to Be Overly Greedy}

\author[1,2]{Xiayang Li}
\author[1,2,*]{Shihua Zhang}

\affil[1]{Academy of Mathematics and Systems Science, Chinese Academy of Sciences}
\affil[2]{School of Mathematics Sciences,
University of Chinese Academy of Sciences}
\affil[*]{\small Corresponding Author}

\begin{document}

\maketitle

\begin{abstract}
Deep learning has shown significant potential in solving combinatorial optimization problems such as the Euclidean traveling salesman problem (TSP). However, most training and test instances for existing TSP algorithms are generated randomly from specific distributions like uniform distribution. This has led to a lack of analysis and understanding of the performance of deep learning algorithms in out-of-distribution (OOD) generalization scenarios, which has a close relationship with the worst-case performance in the combinatorial optimization field. For data-driven algorithms, the statistical properties of randomly generated datasets are critical. This study constructs a statistical measure called \emph{nearest-neighbor density} to verify the asymptotic properties of randomly generated datasets and reveal the greedy behavior of learning-based solvers, i.e., always choosing the nearest neighbor nodes to construct the solution path. Based on this statistical measure, we develop interpretable data augmentation methods that rely on distribution shifts or instance perturbations and validate that the performance of the learning-based solvers degenerates much on such augmented data. Moreover, fine-tuning learning-based solvers with augmented data further enhances their generalization abilities. In short, we decipher the limitations of learning-based TSP solvers tending to be overly greedy, which may have profound implications for AI-empowered combinatorial optimization solvers.
\end{abstract}

\section{Introduction} 
Traveling Salesman Problem (TSP) is a critical and widely applied combinatorial optimization problem. It involves finding the shortest possible route that visits a given set of cities exactly once and returns to the origin city. Due to its NP-hard nature, TSP is a central problem in optimization theory with numerous practical applications in logistics~\cite{baniasadi2020transformation}, routing~\cite{ralphs2003capacitated}, manufacturing~\cite{bagchi2006review}, and even circuit design~\cite{onwubolu2004optimizing}. 

Previous studies have developed various exact and heuristic methods to tackle TSP, including dynamic programming~\cite{held1962dynamic}, branch-and-bound~\cite{volgenant1982branch}, and genetic algorithms~\cite{larranaga1999genetic}. However, due to the combinatorial explosion of possible solutions as the number of cities increases, finding the optimal solution becomes computationally infeasible for large instances. This has led to significant interest in approximation algorithms and learning-based approaches that can offer near-optimal solutions in a reasonable amount of time. Deep learning techniques~\cite{lecun2015deep} provide a novel `data-driven' approach to problem-solving, enabling researchers to address challenges from computational complexity more effectively. The learning-based algorithms can be divided into two main categories~\cite{Jingyan:196322}: i) Learning-based end-to-end algorithms, especially the end-to-end construction algorithm; ii) Hybrid algorithms combining deep learning with heuristic algorithms. This study investigates the potential biases in solving TSP using deep learning solvers. Given the difficulty in isolating the sources of bias in the latter category, this paper focuses on the study of end-to-end construction algorithms.

End-to-end construction algorithms employ neural networks to learn construction heuristics and construct solutions from scratch, exhibiting rapid solving speed. The algorithms typically utilize an encoder-decoder structure. The encoder generates embeddings for nodes, while the decoder computes the probability distribution of candidate nodes based on the information of encoders. Ptr-Net~\cite{vinyals2015pointer} is a typical representative. However, due to the limitations of supervised learning, the solution quality is often low. The algorithms developed based on Ptr-Net~\cite{kool2018attention,Bresson2021,joshi2022learning,yook2024cycleformer,jin2023pointerformer} utilize reinforcement learning algorithms to train models for improving solution quality.

However, due to their data-driven nature, some challenges remain. The existing literature fails to solve TSP instances when training and test data have different distributions. One perspective is the difficulty of scale generalization, which manifests in the larger scale of the test instances. In this paper, we take an orthogonal direction by considering the generalization of algorithms on instances of the same size but having different distributions. Researchers typically train neural network solvers using reinforcement learning methods with random datasets of sizes TSP20/50/100~\cite{Jingyan:196322} generated with node coordinates in the unit square $[0,1]^2$. When tested on a dataset generated with a uniform distribution, the neural solver typically performs well. However, we are unsure whether learning-based solvers would still work well on some structures, which hinders the solver's application in real-world problems. At the same time, analyzing instances of combinatorial optimization problems can effectively identify bottlenecks in learning models and algorithms. With the support of scaling laws~\cite{kaplan2020scalinglawsneurallanguage}, the performance of the large model has improved significantly with size. However, as data resources become increasingly scarce, the importance of data quality in model training is becoming more pronounced~\cite{wu2023dataoptimizationdeeplearning}. Delving into and analyzing combinatorial optimization instances with numerous corner cases will help drive the transformation and development of data-driven approaches. 

In this paper, we analyze the (implicit) statistical properties of datasets generated with the uniform distribution, explain the potential implicit biases in the algorithm called greedy behavior, i.e., always choosing the nearest nodes for the solution construction, and demonstrate that such biases exist. On the other hand, we develop a data augmentation method to improve model robustness. However, for learning-based solvers, can data augmentation truly elevate them to the level of a general-purpose solver~\cite{wang2024asp}? This is one of the questions we aim to answer here.
%\setlength{\fboxsep}{10pt}
%\colorbox{pnasblueback}{\begin{minipage}{0.95\textwidth}
%\textbf{\textcolor{pnasbluetext}{Significance Statement}}
%While deep learning has demonstrated impressive performance in solving combinatorial optimization problems like the traveling salesman problem, its behavior in out-of-distribution scenarios remains poorly understood. Here, we identify that deep learning solvers exhibit a greedy tendency by always selecting the nearest neighbor when constructing solutions when training on RUE instances. Thus, we uncover a key limitation in current learning solvers. Moreover, we propose data augmentation techniques to improve generalization and demonstrate that fine-tuning with augmented data can significantly enhance solver performance. This work highlights the potential directions for improving the ability of AI-empowered combinatorial optimization algorithms.
%\end{minipage}}
\vspace{-10pt} 
\section{Overview}
\begin{figure}[t]
    \centering
    \includegraphics[width=1.0\linewidth]{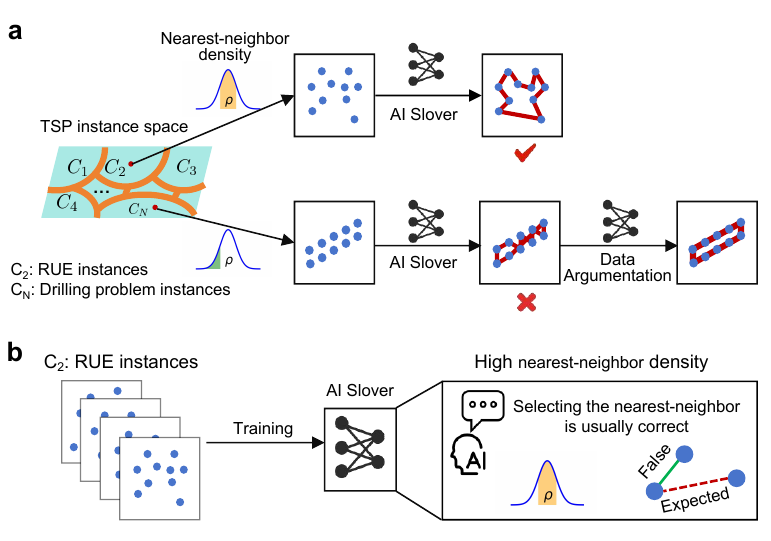}
    \caption{\textbf{(a)} Using random uniform Euclidean (RUE) data for model training may lead to algorithmic bias. In other words, the algorithm cannot guarantee that data from all structures are effective at the same scale. We can explicitly mine instances that cause the algorithm to fail using the nearest-neighbor density measure.
    \textbf{(b)} The mechanism behind the bias introduced by datasets. Specifically, the model is trained on a dataset generated from a uniform distribution, which has a high nearest-neighbor density characteristic. As a result, the model may incorrectly assume a causal relationship between nearest-neighbor density and the solution.}
    \label{fig:Fig1}
\end{figure}

\vspace{-8pt} 
In this study, we construct a statistical measure called the \emph{nearest-neighbor density} and validate the greedy behavior of the learning-based solvers introduced by data sampling from uniform or normal distributions and real-world city instances \textbf{(Fig.~\ref{fig:Fig1}a,b)}. We provide the asymptotic lower bound of the nearest-neighbor density under uniform distribution and formulate a mathematical conjecture. This conjecture could be a significant open problem in probability theory, graph theory, and combinatorial optimization. We leverage the nearest-neighbor density to develop augmented instances for TSP deep-learning solvers. These instances are constructed based on distributional shifts and node perturbations. We verify that these more diverse instances effectively interfere with learning-based solvers. By using these instances as augmented TSP instances, we fine-tune some base solvers~ \cite{joshi2022learning,Bresson2021,kwon2020pomo,jin2023pointerformer,yook2024cycleformer} and confirm a substantial improvement in the generalization performance. Although our data augmentation method based on nearest-neighbor density can improve the learning-based solvers' performance, it is not complete. Unless NP = co-NP, there does not exist an effective and complete data augmentation method. Also, we introduce the concept of efficient algorithm covering and demonstrate that unless NP = P, there is no solution to obtain a general-purpose solver by integrating polynomially many (biased) algorithms. 
\vspace{-5pt} 
\section{The nearest-neighbor density}
\subsection{From the nearest-neighbor algorithm to the nearest-neighbor density}
Given a set of cities $\{c_1,c_2,\cdots, c_N\}$ and  a distances $d(c_i,c_j)$ for each pair of distinct cities $\{c_i,c_j\}$, the goal of TSP is to find an ordering $\pi$ of the cities that minimizes the quantity:
\begin{equation}
\sum\limits_{i=1}^{N-1} d(c_{\pi(i)},c_{\pi(i+1)})+d(c_{\pi(N)}, c_{\pi(1)})
\end{equation}
This quantity is referred to as the tour length that a salesman would make when visiting the cities in the order specified by the permutation, returning to the initial one at the end. We concentrate on the symmetric TSP on the two-dimensional Euclidean plane, satisfying the triangle inequality.

\begin{figure}[htbp]
    \centering
    \includegraphics[width=1.05\linewidth]{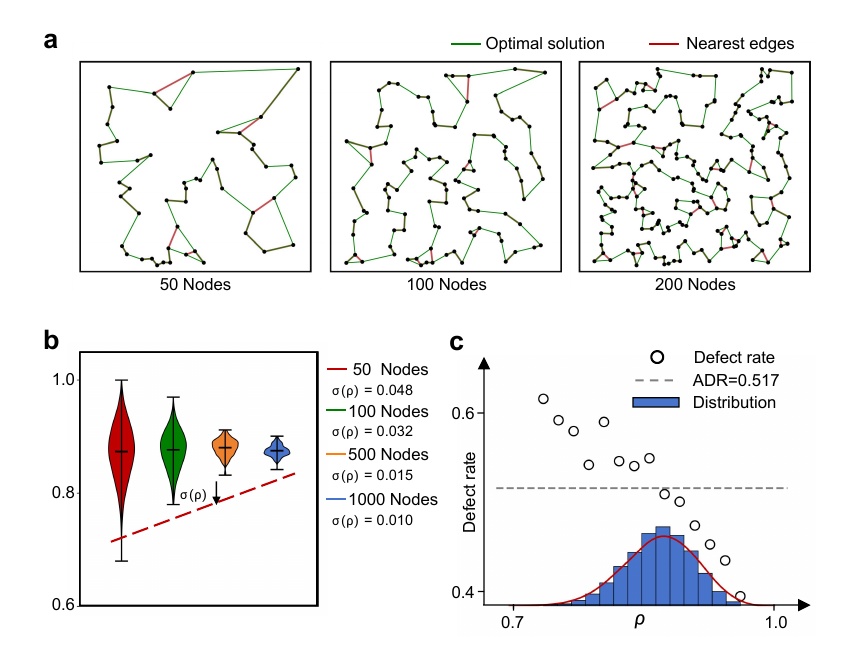}
    \caption{\textbf{(a)} The preference of the model's solutions for nearest-neighbors at different node scales (50/100/200) generated from a uniform distribution. The green line represents the optimal solutions, while the red line represents the nearest-neighbor edges not covered by the optimal solutions. Most of the nearest-neighbor edges are included in the optimal solutions. \textbf{(b)} The statistical characteristics of nearest-neighbor density for instances with different node scales under the uniform distribution. As the node scale increases, the standard deviation decreases. \textbf{(c)} The model's performance on instances with different nearest-neighbor densities under uniform distribution. Generally, the model performs better (worse) on instances with high (low) nearest-neighbor density.} 
    \label{fig:Fig2}
\end{figure}

The most natural heuristic or greedy strategy for TSP is the famous nearest-neighbor algorithm (NN). It always selects the nearest as-yet-unvisited location as the next node. Due to its poor performance, it is generally not used for TSP. There is only a theoretical guarantee that \( \frac{NN(I)}{OPT(I)} \leq 0.5(\lceil \log_2 N\rceil + 1) \), and no substantially better guarantee is possible as there are instances for which ratio grows as $\Theta(\log N)$~\cite{bib1}. In this paper, we consider the question from another view: \textbf{For general instances, what proportion do the nearest-neighbor nodes or edges occupy in the optimal solution?} We generate instances uniformly distributed in a two-dimensional Euclidean plane with node sizes of 50, 100, and 200 and plot the images of the nearest-neighbor edges and the optimal solution (\textbf{Fig.~\ref{fig:Fig2}}). We found that the proportion of nearest-neighbor edges in the optimal solution is very high. For example, in the instance with 50 nodes, only five nearest-neighbor edges do not appear in the optimal solution. 

Here, we propose a statistical measure that characterizes the proportion of nearest-neighbor edges in the optimal solution. We introduce a node-based nearest-neighbor density considering that the algorithm progressively selects the next node to visit during its execution.

\begin{definition}[The Nearest-Neighbor Density]
For a TSP instance \( G=(V,E,d) \), where \( V \) is the set of vertices with the size of \( n \), \( E \) is the set of edges with the size of \( m \) and \( d \) is the weight of the edges, i.e., the Euclidean distance between two points in the two-dimensional Euclidean plane. Given the optimal tour \( \tau^* \) found, we denote $\mathcal{N}(c_i)$ and $\mathcal{N}^\prime(c_i)$ as the set of the nearest-neighbors of $c_i$ on $G$ and on $\tau^*$, respectively. The nearest-neighbor density \( \rho_n^* \) is calculated as follows:
\begin{equation}
    \rho_n^* = \sum\limits_{c_i\in V} \frac{\vert\mathcal{N}(c_i)\bigcap \mathcal{N}^\prime(c_i)\vert}{\vert \mathcal{N}(c_i)\vert\cdot n}.
    \label{eqn1}
\end{equation}
\end{definition}
\begin{remark}
    For a continuous probability distribution sampling dataset, the set of points where a given node $c$ has more than one nearest-neighbor, i.e., two or more nodes are equidistant from $c$ constitute a null set.
    In this case, the calculation of \( \rho_n \) simplifies to a specific form:
    \begin{equation*}
    \rho_n^* = \sum\limits_{c_i\in V} \frac{I(c_i)}{n},
\end{equation*}
where $I(c_i)=\begin{cases} 1 , ~ c_i \text{ is adjacent to } c\in \mathcal{N}(c_i)~on~ \tau^*, \\0, \text{ others}.\end{cases}$ However, in some real-world or synthetic datasets, a node may have multiple nearest neighbors. Therefore, it is necessary to use Equation~\ref{eqn1} for calculation. For specific examples, refer to those provided in TSPLIB.
\label{rem_1}
\end{remark}

\subsection{Calculation of nearest-neighbor density under different distributions}
%Subsequently, we separately calculated the nearest-neighbor density for the random uniform instances, random normal instances, and the benchmark instances from TSPLIB. Among these, random uniform instances are commonly used to training and testing for the traveling salesman problem. The results are as follows.

\subsubsection{Random uniform Euclidean (RUE) dataset}
First, we evaluate the statistical results of the nearest-neighbor density on instances generated from the uniform distribution with the number of nodes varying from 20 to 50 in step 5. The nearest-neighbor density maintains a high level regardless of the node scale, indicating that each node is adjacent to its nearest neighbor on the optimal Hamiltonian circuit with a high probability. We observe an upward trend as the number of nodes increases on the RUE data (\textbf{Table~\ref{tablerho}}).

\begin{table}[h]
  \caption{The estimation of $\rho_n$ for instances generated under different distributions}
  \centering
  \scriptsize
  % \hspace*{-1.3cm}
  \resizebox{0.5\textwidth}{!}{\begin{tabular}{lccc}
    \toprule

 Node Size & \multicolumn{3}{c}{The Nearest Density $\rho_n$}  \\
      & Uniform  & Normal  & Scale-free \\
    \midrule
    20 & 0.8694 & 0.8538 & 0.7596\\
    25 & 0.8685 & 0.8578 & 0.7504\\
    30 & 0.8700 & 0.8601 & 0.7523\\
    35 & 0.8709 & 0.8629 & 0.7493\\
    40 & 0.8722 & 0.8645 & 0.7525\\
    45 & 0.8727 & 0.8665 & 0.7620\\
    50 & 0.8738 & 0.8695 & 0.7571\\
    \bottomrule
  \end{tabular}}
\label{tablerho}
\end{table}

We analyze the aggregation degree of \( \rho_n \) as the number of nodes varies from 50 to 1000 (\textbf{Fig.~\ref{fig:Fig2}b}) and observe that as the node size increases, the standard deviation of \( \rho_n \) decreases, indicating the convergence of $\rho_n$ towards the $\mathbb{E}(\rho_n)$. This suggests that as the node size grows, the representativeness of instances generated by uniform distribution deteriorates.

Intuitively, a larger \( \rho_n \) of an instance indicates a lower difficulty. We test the performance of Transformer-TSP solver~\cite{Bresson2021} on the RUE test set with 10,000 samples of 50 nodes. We classify some samples as `defective' based on the optimal gap and then calculate the defect rate for different instances (\textbf{Fig.~\ref{fig:Fig2}c}). Specifically, we consider the optimal gap (defined as $\frac{\text{NN}(G)}{\text{OPT}(G)}-1$) of the solution output $\text{NN}(G)$ by the Transformer-TSP solver on a specific instance $G$ under the greedy decoding type. When the gap exceeds a threshold (i.e., 0.1\% in this study), the solution is classified as `defective'. The defect rate refers to the proportion of defective solutions for the test instances. Generally, The model performs better (worse) on instances with high (low) nearest-neighbor density.

Based on the above experiments, we analyze the asymptotic bounds of the nearest-neighbor density \( \rho_n \) when $n\rightarrow \infty$.
\begin{theorem}
Given the TSP instance \( G=(V,E,b) \) on the two-dimensional Euclidean plane and \( c_i=(x_i,y_i)\in V(G) \), where \( (c_i)_{i=1}^{|V|} \sim \text{Uniform}([0,1]\times [0,1]) \), the nearest-neighbor density \( \rho_n \) has an asymptotic lower bound $\rho_n \geq \frac{27-32\beta}{7}$, as \( n \rightarrow \infty \), a.s..
\end{theorem}
\begin{remark}
    $\beta$ is a known constant called the \textbf{Euclidean TSP Constant}. When placing $n$ cities on a square of area $[0,1]\times [0,1]$. uniformly at random, the optimal tour length $l_{opt}$ approaches the limit~\cite{Beardwood59}:
    \begin{equation}
        \lim\limits_{n\rightarrow \infty}\frac{l_{\text{opt}}}{\sqrt{n}} = \beta.~\text{a.s.}
    \end{equation}
    The best estimate for $\beta$ is $0.7124 $~\cite{bib2}. Thus our lower bound for $\rho_n$ takes $0.6005$ approximately. However, the theoretical upper bound estimation of the TSP constant for TSP remains an open problem. Currently, the best theoretical upper bound obtained is $\beta\leq 0.90304$.
    \label{remark1}
\end{remark}
According to our observation in \textbf{Table~\ref{tablerho}}, we propose the following conjecture.
\begin{conjecture}
    Given the TSP instance \( G=(V,E,b) \) on the two-dimensional Euclidean plane and \( c_i=(x_i,y_i)\in V(G) \), where \( (c_i)_{i=1}^{|V|} \sim \text{Uniform}([0,1]\times [0,1]) \), the following two statements hold 
    \begin{enumerate}
        \item $\mathbb{E}(\rho_n)$ increases with $n$.
        \item $\rho_n\rightarrow_{p} \mathbb{E}(\rho_n), \text{ as } n\rightarrow \infty$, where $\mathbb{E}(\rho_n)>0.87, n\rightarrow \infty$.
    \end{enumerate}
\label{conj} 
\end{conjecture}

We can show that as the node size increases, the instances obtained from the uniform distribution probabilistic sampler become less representative (as reflected in the decreasing standard deviation of \( \rho_n \) with increasing node size (\textbf{Fig.~\ref{fig:Fig2}}b). Moreover, this creates a simpler subproblem: when we deal with random uniform Euclidean instances, the greedy algorithm is not as bad as imagined, which is counterintuitive.

\subsubsection{Random normal Euclidean (RNE) dataset}
We also estimate $\rho_n$ of normal Euclidean instances with the number of nodes varying between 20 and 50. According to \textbf{Table ~\ref{tablerho}}, the nearest-neighbor density remains in a relatively high level ($\geq 0.85$).

\subsubsection{TSPLib95 dataset}
TSPLib95 is a well-known benchmark library\cite{reinelt1995tsplib95} of sample instances for TSP and related combinatorial optimization problems. In this part, we selected 38 instances from TSPLib95 (with node sizes less than 400 for computational convenience and the problem type is \texttt{EUC\_2D}) to test their nearest-neighbor density (\textbf{Table~\ref{table:Real}}). The results show that some datasets have a nearest-neighbor density smaller than 0.85, indicating that the topological characteristics of the city datasets may not be approximated well by simple distributions. Notably, the $\rho_n$ values for the two instances, \texttt{a280} and \texttt{pr107}, are less than 0.8. We will discuss the potential patterns in such instances later.

\begin{table}[h]
    \caption{Real TSP datasets}
    \centering
    \begin{tabular}{ccc|ccc}
        \toprule
        Name & Node Size & $\rho_n$ & Name& Node Size & $\rho_n$ \\
        \midrule
        eil51 & 51 & 0.8529 & pr144 & 144 & 0.9583\\
        berlin52 & 52 & 0.8077 & ch150 & 150 & 0.8800\\
        st70 & 70 & 0.8286 & pr152 & 152 & 0.9342 \\
        eil76 & 76 & 0.8092 & u159 & 159 & 0.9444\\
        pr76 & 76 & 0.9211 & rat195 & 195 & 0.9359\\
        rat99 & 99 & 0.8485 & d198 & 198 & 0.8283\\
        KroA100 & 100 & 0.8700 & KroA200 & 200 & 0.8550\\
        KroB100 & 100 &0.8700 & KroB200 & 200 & 0.9200\\
        KroC100 & 100 & 0.9100 & ts225 & 225 & 0.8267\\
        KroD100 & 100 & 0.8600 & tsp225 & 225 & 0.8689\\
        KroE100 & 100 & 0.9100 & pr226 & 226 & 0.8842\\
        rd100 & 100 & 0.8300 & gil262 & 262 & 0.8779\\
        eil101 & 101 & 0.8218 & gil262 & 262 & 0.8798\\
        lin105 & 105 & 0.9238 & pr264 & 264 & 0.9564\\
        pr107 & 107 & 0.7555 & a280 & 280 & 0.7952\\
        pr124 & 124 & 0.9556 & pr299 & 299 & 0.8434\\
        bier127 & 127 & 0.8674 & pr299 & 299 & 0.8434\\
        ch130 & 130 & 0.8615 & lin318 & 318 & 0.9025\\
        pr136 & 136 & 0.8161 & rd400 & 400 & 0.8700\\
        \bottomrule
    \end{tabular}
    \label{table:Real}
\end{table}
\section{Data augmentation}

\subsection{TSP instances induced by scale-free network}
\begin{figure}[t]
    \centering
    \includegraphics[width=1.01\linewidth]{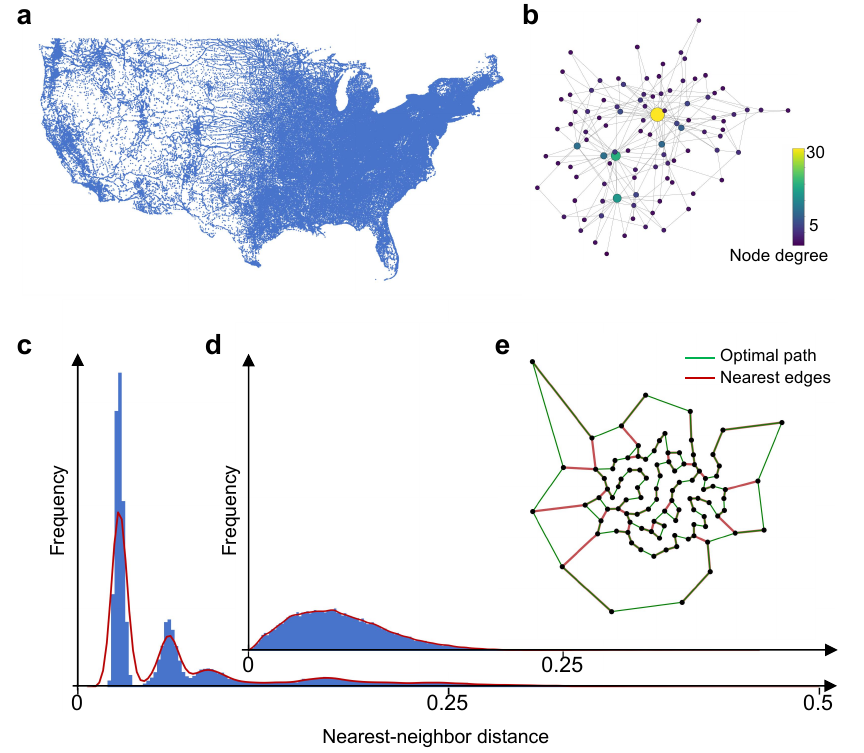}
    \caption{\textbf{(a)} Illustration of a TSP instance containing 115,475 towns in the United States from the USA TSP Challenge. 
    \textbf{(b)} A scale-free network with 100 nodes generated using the Barabási-Albert model (Appendix ~\ref{Appendix-BA}). \textbf{(c)} Distribution of nearest-neighbor distances for the TSP instances generated using Algorithm 1, exhibiting a multimodal distribution. \textbf{(d)} Distribution of nearest-neighbor distances for the traveling salesman problem instances generated using the uniform distribution. \textbf{(e)} Visualization of the instance with a size of 100 generated by Algorithm 1.}
    \label{fig:Fig3}
\end{figure}

Scale-free networks~\cite{barabasi1999emergence} are a type of complex network characterized by a small number of highly connected nodes and many nodes with fewer connections. This structure follows a power-law degree distribution, i.e., $P(d)\propto d^{-\alpha}$. Scale-free networks are commonly found in various real-world systems, such as social networks, the internet, and biological networks~\cite{barabasi2009scale}. 
For the TSP, the topological structure of the nodes is trivial, and the metric information between nodes is critical. The nearest-neighbor distance is closely related to the nearest-neighbor density we proposed, and the nearest-neighbor distance reflects the local clustering feature of nodes. For example, in reality, the clustering feature in developed and underdeveloped areas is different, such as in the distribution map of town nodes in the United States (\textbf{Fig.~\ref{fig:Fig3}a}). However, for the uniform euclidean distribution with node size $n$, the probability density function of the nearest-neighbor distance $P(r_1)\propto r (1-\pi r^2)^{n-2}$, so the tail of the nearest-neighbor distance distribution decays rapidly especially when $n$ is large (\textbf{Fig.~\ref{fig:Fig3}d}).

In this paper, we first generate a scale-free network using the Barabási-Albert model~\cite{albert2002statistical}, then convert the degree features of the nodes into distance features. Specifically, it is manifested as points with higher nodes exerting a stronger attraction on neighboring nodes. Next, we use a spring layout algorithm~\cite{fruchterman1991graph} to restore the node coordinates based on the distance information. Details can be found in \textbf{Algorithm~\ref{alg-1}}. Thus, the greater the node degree, the stronger its attraction, resulting in a radial distribution of nodes from the center to the periphery. In particular, we found that the nearest-neighbor distances follow a multimodal distribution in our case.

This indicates that the distribution of nodes exhibits a multilevel morphology in such cases with nodes closer to the center being denser and those closer to the edges being sparser. When addressing the TSP in this context, the model needs to be more \emph{sophisticated} and cannot solely rely on nearest-neighbor distances, especially during hierarchical transitions from sparse regions to dense regions. As usual, we calculated the nearest-neighbor density $\rho_n$ on this type of instances and found that it is lower compared to the instances discussed in the previous section (\textbf{Table~\ref{tablerho}}).

\subsection{Drilling problem instances}
\begin{figure}[!ht]
    \centering
    \includegraphics[width=0.95\linewidth]{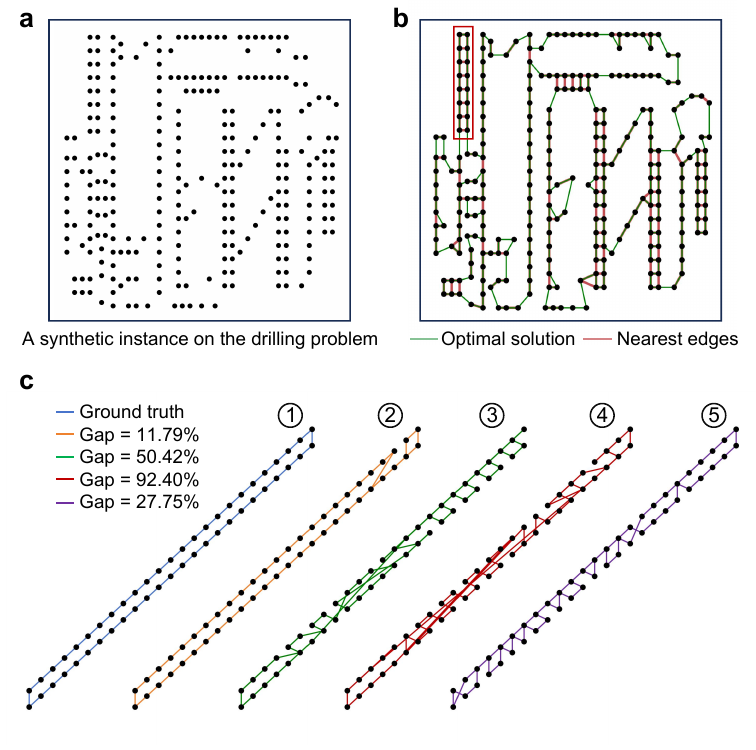}
    \caption{
\textbf{(a)} Visualization of the a280 instance from TSPLIB. \textbf{(b)} The optimal solution derived from the a280 instance, where the green line represents the optimal solution, the red line represents the nearest-neighbor edges not covered by the optimal solution, and the orange highlights the parallel pattern layout observed in this type of instance. \textbf{(c)} The artificial instances constructed based on the pattern in the drilling problem with \textcircled{1} representing the exact solution and \textcircled{2}-\textcircled{5} indicating the solutions of TSP-GNN~\cite{joshi2022learning}, TSP-Transformer~\cite{Bresson2021}, CycleFormer~\cite{yook2024cycleformer}, and the large-scale version of CycleFormer.}
    \label{fig:Fig4}
\end{figure}

The drilling problem~\cite{onwubolu2004optimizing} is an industrial optimization task to model the optimal path for drilling predetermined locations in mechanical manufacturing or electronic device production (e.g., circuit boards or mechanical parts). The objective is to minimize the distance or time for the drill head to move between drilling locations. These points often follow specific distribution patterns, reflecting practical production requirements and potentially displaying some regularity. Compared to general TSP instances sampled by the uniform distribution, where node layouts might be random or lack a specific arrangement, the points in the drilling problem are often associated with specific processes or product structures (e.g., \texttt{a280.tsp} in the TSPLIB dataset) (\textbf{Fig.~\ref{fig:Fig4}a,b}).

\begin{algorithm}[t]
    \caption{TSP-Instances Induced by Scale-Free Network}
    \begin{algorithmic}[1]
        \STATE \textbf{Input:} Parameters for \texttt{Barabási-Albert model}, tunable constant \(k\), the instance size $n$.
        \STATE Use the \texttt{Barabási-Albert model} to generate a scale-free network.
        \FOR{each pair of nodes \(u\) and \(v\) in the network}
            \STATE Let \(d_u\) and \(d_v\) be the degrees of nodes \(u\) and \(v\) respectively.
            \STATE Calculate the distance \(d_{uv}\) between nodes \(u\) and \(v\): 
             \(d_{uv} = \exp(-k*(\deg (u) +\deg (v)))\) where \(k\) is a tunable positive constant.
        \ENDFOR
        \STATE Use the \textbf{\texttt{spring layout algorithm}} to restore the position coordinates of the nodes based on distance information.
        \STATE \textbf{Output: }Obtain TSP instances where the nearest-neighbor distance follows a power law distribution.\\
    \end{algorithmic}
    \label{alg-1}
\end{algorithm}

It is easy to see that the nodes in the drilling problem instance exhibit a locally parallel grid-like layout, which often results in parallel line segments in the tour path (\textbf{Fig.~\ref{fig:Fig4}a}). Similar examples are also frequently found in real-world city-based datasets. From the arrangement of shops along parallel streets to city layouts shaped by specific terrain or historical reasons, there can often be a relatively more regular grid-like structure.

Instances with grid-like patterns may seem relatively easy to address. If the spacing between two parallel grid segments is smaller than the spacing between points within each segment, the nearest-neighbor density will significantly decrease. We construct synthetic instances to demonstrate it. Specifically, we sample \(\frac{n}{2}\) equally spaced points on each of the parallel lines \(y = x\) and \(y = x + 0.05\), where \(n\) is the size of the TSP instance. It is straightforward to calculate that the nearest-neighbor density on this constructed counterexample is \(\rho_n = \frac{2}{n}\) when $n$ is sufficiently large. We used several deep learning-based solvers on this type of data and found that the results were unsatisfactory (\textbf{Fig.~\ref{fig:Fig4}c}).

\subsection{Fine-tune the base solver with augmented instances}
\begin{figure}[h]
    \centering
    \includegraphics[width=0.95\linewidth]{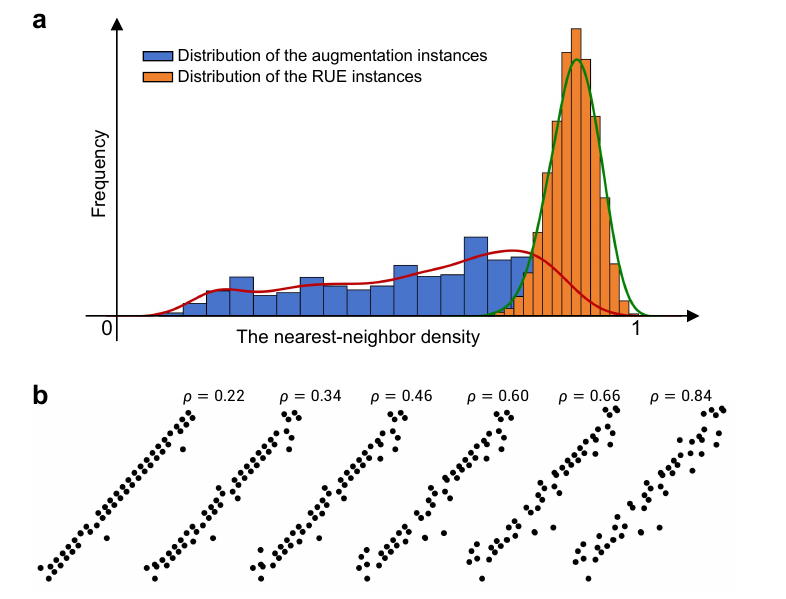}
    \caption{
\textbf{(a)} Distribution of the nearest-neighbor density \( \rho_{50} \) for the augmented instances constructed by perturbing the drilling model instances. \textbf{(b)} Visualization of nine augmented instances with \( \rho_{50} \) ranging from 0.1 to 0.9.}
    \label{fig:Fig5}
\end{figure}

Utilizing the examples from the previous subsection as the basic construction instances, we create a perturbation-based instance augmentation algorithm and generate an instance set of 10000 sample instances with a node size of 50, where \(\rho_{50}\) spans values between 0.06 and 0.90 (\textbf{Fig.~\ref{fig:Fig5}b}). The core idea is similar to stochastic offset data generation, where Gaussian perturbations of different scales are applied to the basic construction instances. The process generates instances along the transformation path from basic construction instances to RNE instances, which in turn results in samples with more diverse nearest neighbor density distributions ($\textbf{Algorithm}~\ref{alg-2}$).
\begin{algorithm}
\caption{Generate Perturbed Parallel Line Points with Rotation}
\begin{algorithmic}[1]
\STATE\textbf{Input:} Total number of points $n$, large noise ratio $\alpha$, small noise ratio $\beta$,
\STATE \hspace{2.2em} large noise standard deviation $\sigma_{large}$, small noise standard deviation $\sigma_{small}$
\STATE \textbf{Output:} Set of perturbed points

\STATE Randomly select a slope $s$ and a distance $d$ between two parallel lines and define two parallel lines on the $[0, 1] \times [0, 1]$ plane using $s$ and $d$

\STATE Generate $n/2$ equally spaced points on each line segment, and combine points from both lines to form a set of $n$ points, denoted as $\mathcal{P}$
\STATE Randomly select $\alpha \cdot n$ points from $\mathcal{P}$ and apply large Gaussian noise with standard deviation $\sigma_{large}$ and apply small Gaussian noise with standard deviation $\sigma_{small}$ to the remaining $\beta \cdot n$ points

\STATE Uniformly sample a rotation angle $\theta$ from $[0, 180]$ degrees and rotate all points in $\mathcal{P}$ by angle $\theta$

\STATE \textbf{Return} the final set of rotated and perturbed $\mathcal{P}$.
\end{algorithmic}
\label{alg-2}
\end{algorithm}

We performed full fine-tuning on several classical neural solvers using augmented data (\textbf{Table~\ref{Table3}}). It is worth noting that, without fine-tuning, these neural solvers showed varying degrees of performance decline on the constructed instances, with CycleFormer~\cite{yook2024cycleformer} performing the worst. Compared with the base solvers, most solvers show performance improvements on the constructed instances such as the drilling model and scale-free model instances. Additionally, we observed that except for POMO, other neural solvers also achieved improvements on standard benchmarks (e.g., the RUE instances) after fine-tuning (The experimental details could refer to Appendix ~\ref{appendix-ed}). 

To further explore the generality of our data augmentation method, we also tested the performance of solvers before and after fine-tuning on the synthetic instances presented in the previous work~\cite{wang2024asp}, which adopt the convolution distribution of both uniform and normal distributions, resulting in varying degrees of performance degradation for all solvers. However, after fine-tuning, we observed performance improvements for all solvers except TSP-GNN, with Transformer-TSP showing particularly significant gains—achieving performance improvements of 86\% and 93\% under the greedy decoding and beam search decoding concerning the optimal gap, respectively. This indicates that fine-tuning with augmented data enables solvers to capture essential features better when solving TSP.

Moreover, the experiments reveal that CycleFormer may suffer from severe overfitting issues in the RUE instances. Its performance was the worst among all solvers on drilling model and convolution distribution instances, though CycleFormer performs exceptionally well in the RUE instances. Even after fine-tuning, it failed to achieve notable improvements in the convolution distribution instances. We also observed that before fine-tuning, the larger version of CycleFormer outperforms the original CycleFormer on the RUE instances but performs much worse on the drilling model. This indicates that simply pursuing an increase in model size while ignoring data quantity or quality is unscientific, and ignoring dataset bias or solely focusing on improving benchmark performance during neural solver training can lead to serious consequences that may not be hard to detect in ideal experimental environments.

\begin{table}
  \caption{Results on the RUE, scale-free, drilling model and convolution distribution TSP Instances}
  \label{Table3}
  \centering
  \scriptsize
    \hspace*{-0.5cm}
  \begin{tabular}{lccccccccc}
    \toprule
    % \multicolumn{2}{c}{Part}                   \\
    % \cmidrule(r){1-2}
    \textbf{Algorithm} & \textbf{Decoding} & \multicolumn{2}{c}{\textbf{RUE }} & \multicolumn{2}{c}{\textbf{Scale-free Model }} & \multicolumn{2}{c}{\textbf{Drilling Model }} & \multicolumn{2}{c}{\textbf{Convolution Distribution}}\\
     & \textbf{Type} & Length  & Gap (\%)  & Length  & Gap (\%) & Length  & Gap (\%) & Length  & Gap (\%)\\

    \midrule
    CONCORDE~\cite{applegate1995finding,applegate1998solution} & Exact & 5.688 & 0 & 4.705 & 0 & 1.965 & 0  & 1.921 & 0 \\
    \midrule
    TSP-GNN~\cite{joshi2022learning} & Greedy   & 5.928    &  4.227   &  5.375   & 14.238  & 2.769    & 40.945 & 2.290 & 19.189 \\
    \rowcolor{pnasblueback} TSP-GNN (+DA)  & Greedy   & \textbf{5.924}    &  \textbf{4.157}   &  \textbf{5.302}  & \textbf{12.687}  & \textbf{2.574}    & \textbf{31.020} & \textbf{2.260} & \textbf{17.627} \\
    TSP-Transformer~\cite{Bresson2021} & Greedy & 5.702 & 0.248 & 4.844&  2.968 & 2.540 & 29.838  & 2.495 & 30.730\\
    \rowcolor{pnasblueback} TSP-Transformer (+DA) & Greedy & \textbf{5.699} & \textbf{0.192} & \textbf{4.803} & \textbf{2.095} & \textbf{1.977} & \textbf{0.637} &\textbf{2.002} & \textbf{4.280} \\
    POMO~\cite{kwon2020pomo} & Greedy & 5.708 & 0.365 & 4.994  & 6.143 & 2.161 & 9.992  & 1.988 & 3.460 \\
    \rowcolor{pnasblueback} POMO (+DA) & Greedy & 5.710 & 0.385 & 5.011  & 6.506 & \textbf{2.005} & \textbf{2.037}  & \textbf{1.988} & \textbf{3.445} \\
    Pointerformer~\cite{jin2023pointerformer} & Greedy & 5.709 &  0.371 & 5.044  & 7.212 & 2.572 & 30.933 & 2.396 & 24.722 \\
    \rowcolor{pnasblueback} Pointerformer (+DA) & Greedy & \textbf{5.707} &  \textbf{0.347} & \textbf{5.036}  & \textbf{7.031} & \textbf{1.988} & \textbf{1.166} & \textbf{2.335} & \textbf{21.547} \\
    CylceFormer~\cite{yook2024cycleformer}& Greedy &5.723 &0.593  &5.317 & 13.008 & 3.344 & 70.228 & 3.319 & 72.752 \\
    \rowcolor{pnasblueback} CycleFormer (+DA) & Greedy & \textbf{5.720} & \textbf{0.563} & \textbf{5.210} & \textbf{10.744} & \textbf{2.065} & \textbf{1.965} & \textbf{3.213} & \textbf{66.753}\\
    CycleFormer,Large~\cite{yook2024cycleformer} & Greedy & 5.699  & 0.207 &5.156  & 9.592 & 3.797 & 93.356 & 3.085 & 60.573 \\   
    \rowcolor{pnasblueback} CycleFormer,Large (+DA) & Greedy & \textbf{5.696}  & \textbf{0.154}  & \textbf{5.131}  & \textbf{9.048} & \textbf{1.977} & \textbf{0.643} & \textbf{2.589} & \textbf{32.806} \\
    \midrule
    TSP-GNN~\cite{joshi2022learning} & Beam search   &  5.767   & 1.396    &  5.062  &  7.586    &   2.559  & 30.256 & 2.089 & 8.727\\
    \rowcolor{pnasblueback} TSP-GNN (+DA) & Beam search   &  \textbf{5.766}   & \textbf{1.379}    &  \textbf{5.044}  &  \textbf{7.203}    &   \textbf{2.404}  & \textbf{22.366} & 2.102 & 9.404 \\
    TSP-Transformer~\cite{Bresson2021} & Beam search & 5.689 & 0.027 &  4.756 & 1.089 & 2.336&  19.322 & 2.382 & 24.867 \\
    \rowcolor{pnasblueback} TSP-Transformer (+DA) & Beam search & \textbf{5.689} & \textbf{0.018} & \textbf{4.735} & \textbf{0.657} & \textbf{1.966} & \textbf{0.061} & \textbf{1.953} & \textbf{1.673}\\
    POMO~\cite{kwon2020pomo} & Beam search & 5.694 & 0.116 & 4.824 & 2.536 & 2.080 & 5.875 & 1.949 & 1.451\\
    \rowcolor{pnasblueback} POMO (+DA) & Beam search & 5.694 & 0.118 & 4.836 & 2.795 & \textbf{1.980} & \textbf{0.789} & \textbf{1.946} & \textbf{1.279}  \\
    Pointerformer~\cite{jin2023pointerformer}& Beam search & 5.693 & 0.100 & 4.807 & 2.160 & 2.226 & 13.326 & 2.177 & 13.292 \\
    \rowcolor{pnasblueback} Pointerformer (+DA) & Beam search & 5.694 & 0.109 & \textbf{4.783} & \textbf{1.654} & \textbf{1.967} & \textbf{0.112} & \textbf{2.095} & \textbf{9.024} \\
    CylceFormer~\cite{yook2024cycleformer}& Beam search & 5.691 & 0.056 &4.793 & 1.864 & 2.604 & 32.531 & 2.980 & 55.127 \\
    \rowcolor{pnasblueback} CycleFormer (+DA) & Beam search & \textbf{5.690} & \textbf{0.039} & \textbf{4.761} & \textbf{1.193} & \textbf{1.973} & \textbf{0.453} & \textbf{2.878} & \textbf{49.409}\\
    CycleFormer,Large~\cite{yook2024cycleformer} & Beam search &  5.689 & 0.032 & 4.763  &1.233  & 2.946 & 49.938 & 2.749 & 43.062 \\
    \rowcolor{pnasblueback} CycleFormer,Large (+DA) & Beam search &  5.690 & 0.042 & \textbf{4.743}  & \textbf{0.800}  & \textbf{1.966} & \textbf{0.083} & \textbf{2.284} & \textbf{18.542} \\
    \bottomrule
  \end{tabular}

  % 注释部分
  \vspace{0.5em}
  \raggedright
  \textit{Note}: All gaps (\%) are calculated relative to the results from CONCORDE as the baseline. (+DA) refers to the base solvers fine-tuned on augmented data proposed by us. \textbf{Bold} text indicates performance gains after fine-tuning on the current instance set.
\label{Tab:3} 
\end{table}

\section{Limitations of data-driven methods: Is augmentation all you need?}

We have two problems to solve in this section. 
\begin{enumerate}
    \item \textbf{(About Data)} As mentioned earlier, we constructed two types of instances with low nearest-neighbor density, enhancing the robustness of the network as training data. However, the examples we constructed exhibit distinct characteristics. Are there other instances that are not covered by our construction? 
    \item \textbf{(About Model)} On uniform distribution, we seem to be able to obtain effective algorithms using neural network methods. However, this efficient algorithm for uniform distribution cannot cover all instances. Can we utilize different algorithms, i.e., ensemble methods, to cover all instances? 
\end{enumerate} 
In the following, we present some (negative) results regarding the following two questions separately.

\subsection{No efficient complete generator based on $\rho$}
First, we need to clarify whether there are more general augmentation methods based on the nearest-neighbor density $\rho_n$, such as those using generative deep learning. Unfortunately, we will prove that any polynomial-time generative method is not complete. As a preparation, we first introduce several concepts of computational complexity~\cite{yehuda2020s,sanchis1990complexity}.
\begin{definition}[Language]
A language L is a set of strings. Every language L induces a binary classification task: the positive class contains all the strings in L, and the negative class contains all the strings in L’s complement $L^C$.
\end{definition}
\begin{definition}[Generator]
A generator for a language $L$ is a nondeterministic Turing machine. A generator is \textbf{complete} if it can generate every example: for every sufficiently large $n$, for every $w\in L$ of length $n$ it holds that $w\in S_L(n)$. A generator $S_L$ is called \textbf{efficient} if it runs in polynomial time.
\end{definition}

\begin{remark}
    The definition of generator implies generating both a sample $w$ and its correct label $(w\in L)$. One widely used efficient generator satisfies this definition which starts with a seed set of deterministically-labeled samples and applies class-preserving rewrites.
\end{remark}
The following theorem indicates that for a generator for TSP instances based on $\rho_n$, only one of efficiency and completeness can be achieved.
\begin{theorem}
    There is no efficient complete generator to generate the instances with the nearest-neighbor density $\rho_n \leq \delta$, where $\delta\in (0,1)$, unless NP = CoNP.
    \label{theorem_3}
\end{theorem}

As mentioned before, we can cover the range of nearest-neighbor density values by using \textbf{Algorithm~\ref{alg-2}}. However, according to \textbf{Theorem~\ref{theorem_3}}, since \textbf{Algorithm~\ref{alg-2}} operates in polynomial time,  it is still not complete. In other words, for a specific nearest-neighbor density value, there exist certain combinatorial graph structures that cannot be generated by \textbf{Algorithm~\ref{alg-2}}. Moreover, there will not exist an efficient and complete sampler for nearest-neighbor density $\rho$, such as generating instances via neural networks. 

\subsection{No efficient algorithmic coverage}
We have found that existing neural network algorithms perform excellently on the RUE TSP instances, at least when the problem scale is not particularly large. In this paper, we discover and confirm that learning-based solvers are biased. Nevertheless, is it possible to train solvers with different preferences for different problem scenarios and use ensemble methods to obtain a universal solver?  We will explain that achieving this would at least require an exponential number of solvers in the ensemble unless P = NP. 
\begin{definition}[Efficient Algorithmic Coverage]
For a combinatorial optimization problem \( \mathcal{P} \in \text{NPC}\), let \( n \) be the size of the instances, \( C \) be the set of all problem instances of size \( n \), $C_1, C_2, \cdots, C_N$ are  subsets of $C$,and $N=O(\text{poly}(n))$ . For every $C_i$, there exists a corresponding efficient algorithm \( \mathcal{A}_i \). If $\{C_i, \mathcal{A}_i\}_{i=1}^N$ satisfies the following properties: 
\begin{itemize}
    \item (Completeness) \( \bigcup_{i=1}^N C_i = C \) ensures that all instances are represented.
    \item (Effectiveness) $\forall x_i\in C_i$, $\frac{\vert \mathcal{A}(x_i)-\mathcal{A}^*(x_i)\vert}{A^*(x_i)}\leq\epsilon$.
    \item (Independence) For all \( C_i \), there exists \( \tilde{x}_i \in C_i \), such that \[ \forall j \neq i, \frac{|\mathcal{A}_j(\tilde{x}_i) - \mathcal{A}^*(\tilde{x}_i)|}{|\mathcal{A}^*(\tilde{x}_i)|} > \epsilon, \] where $
    \mathcal{A}^*$ is the exact algorithm and $\varepsilon$ is the approximation ratio.
\end{itemize}
then we denote \( \{(C_i, \mathcal{A}_i)\}_{i=1}^{N} \) as the efficient algorithmic coverage for \( \mathcal{P} \) with instance size \( n \).
\end{definition} 
\begin{theorem}
There does not exist an exact efficient algorithmic coverage for $\mathcal{P}\in\text{NPC}$ unless \( \text{NP} = \text{P} \).
\end{theorem}

\begin{theorem}
    If the original problem $\mathcal{P}\in \text{NPC}$ does not have a polynomial-time approximation algorithm with the approximation ratio of $\varepsilon$, there does not exist an efficient algorithmic coverage with the approximation ratio of $\varepsilon$  for $\mathcal{P}$.
\end{theorem}

The above results are consistent with our previous discussion. From \textbf{Figure~\ref{fig:Fig2}} and \textbf{Conjecture~\ref{conj}}, we know that the concentration of nearest-neighbor density in the uniform distribution instances increases as the number of nodes grows. This means that, from the perspective of nearest-neighbor density, the range of scenarios that the uniform distribution can cover becomes increasingly limited. Therefore, as the number of nodes increases, the number of specific solvers required for different instances will also grow. The results indicate that this growth is at least exponential.

\section{Related works}
\textbf{The generalization ability of neural solvers.} Existing learning-based solvers often struggle with generalization when faced with changes in problem distributions. Some studies have focused on creating new distributions~\cite{wang2021game, zhang2022learning}. Zhang \emph{et al.}~\cite{zhang2022learning} defined the concept of hardness of TSP instance. However, the definition of hardness is solver-specific. We know this type of instance is difficult for some solvers, but do not understand why. Min \emph{et al.} evaluated the solver-agnostic hardness of various distributions based on the parameter $\tau=l_{opt}/\sqrt{nA}$ defined in the reference~\cite{gent1996tsp}, where $A$ denotes the area covered by the TSP instance, $l_{opt}$ represents the length of the optimal solution and $n$ is the number of cities. However, this complexity is only applicable to decision TSP, whose goal is to determine whether there exists a path whose total distance does not exceed a given threshold. Lischka \emph{et al.}\cite{lischka2024less} used different data distributions to train and test their solvers, but their data augmentation is based on the traditional mutation operators~\cite{bossek2019evolving}. They did not show why the learning solvers training on the uniform distribution dataset may fail. In this paper, we address this challenge by defining a statistic called \emph{the nearest-neighbor density} that not only reflects the inherent learning difficulty of the samples but is also closely related to the solving process of prediction-based learning algorithms. We believe it can better capture the greedy behavior in the learning-based solvers introduced by the training data.

\textbf{The design of universal neural solvers.} Can deep learning learn to solve any task? Yehuda \emph{et al.} \cite{yehuda2020s} showed that any polynomial-time data generator for an NP-hard classification task will output data from an easier NP $\bigcap$ coNP task. However, their theoretical results only apply to data generators that provide labels. Methods that do not require labels, such as reinforcement learning, do not suffer from the problem above. Can we use reinforcement learning to train a universal solver? Wang \emph{et al.}~\cite{wang2024asp} proposed an approach to address the generalization issues and provided a so-called `universal solver' applied to a wide range of problems. However, they do not guarantee the so-called universality. A wider range of applications does not necessarily imply universality. Unfortunately, we will demonstrate that such an approach cannot yield a universal solver. The other potential solution to the generalization problem is to train different solvers in an ensemble manner~\cite{gil2023evolving}. However, we will demonstrate that such an approach is computationally intensive (non-polynomial).

\section{Conclusion}
We introduced the nearest-neighbor density to describe the nearest-neighbor bias introduced by the training dataset (uniform distribution) in neural solvers for TSP. Extensive experiments thoroughly validated this. In particular, \textbf{conjecture 2} may serve as an intriguing open problem in the intersection of combinatorial optimization and probability theory.

We developed a data augmentation method based on nearest-neighbor density and real-world scenarios to alleviate the nearest-neighbor preference. Fine-tuning the base solver with these augmented instances significantly improved its generalization performance. However, this does not imply that we have achieved a universal solver. Data augmentation as a solution is not fundamental to address generalization challenges in combinatorial optimization problems. We analyze the limitations of data-driven methods, highlighting that achieving a universal neural solver through data augmentation (either manual or adaptive) or ensemble methods is infeasible, even for small-scale instances. 

The challenge faced by deep learning models lies in handling corner cases, which is precisely a characteristic of combinatorial optimization problems. Therefore, pursuing a universal neural solver seems unrealistic. Focusing on real-world optimization tasks might be more meaningful than improving performance on unrepresentative benchmarks. Developing fair and appropriate benchmarks to evaluate neural solvers is one of our future research directions. 

On the other hand, due to their lack of interpretability, neural networks are prone to shortcut learning—solving complex problems by capturing superficial features in the data. How to better design data representations or model architectures to reduce the network's reliance on superficial features is another critical question for the fields of AI for combinatorial optimization or even AI for Science.

\bibliographystyle{plain}
\bibliography{neurips_2024}

\begin{thebibliography}{10}

\bibitem{albert2002statistical}
R{\'e}ka Albert and Albert-L{\'a}szl{\'o} Barab{\'a}si.
\newblock Statistical mechanics of complex networks.
\newblock {\em Reviews of modern physics}, 74(1):47, 2002.

\bibitem{applegate1995finding}
David Applegate, Robert Bixby, Va{\v{s}}ek Chv{\'a}tal, and William Cook.
\newblock {\em Finding cuts in the TSP (A preliminary report)}, volume~95.
\newblock Citeseer, 1995.

\bibitem{applegate1998solution}
David Applegate, Robert Bixby, William Cook, and Vasek Chv{\'a}tal.
\newblock On the solution of traveling salesman problems.
\newblock 1998.

\bibitem{bagchi2006review}
Tapan~P Bagchi, Jatinder~ND Gupta, and Chelliah Sriskandarajah.
\newblock A review of tsp based approaches for flowshop scheduling.
\newblock {\em European Journal of Operational Research}, 169(3):816--854, 2006.

\bibitem{baniasadi2020transformation}
Pouya Baniasadi, Mehdi Foumani, Kate Smith-Miles, and Vladimir Ejov.
\newblock A transformation technique for the clustered generalized traveling salesman problem with applications to logistics.
\newblock {\em European Journal of Operational Research}, 285(2):444--457, 2020.

\bibitem{barabasi2009scale}
Albert-L{\'a}szl{\'o} Barab{\'a}si.
\newblock Scale-free networks: a decade and beyond.
\newblock {\em science}, 325(5939):412--413, 2009.

\bibitem{barabasi1999emergence}
Albert-L{\'a}szl{\'o} Barab{\'a}si and R{\'e}ka Albert.
\newblock Emergence of scaling in random networks.
\newblock {\em science}, 286(5439):509--512, 1999.

\bibitem{Beardwood59}
John H.~Halton Beardwood~Jillian and John~Michael Hammersley.
\newblock The shortest path through many points.
\newblock {\em Mathematical proceedings of the Cambridge philosophical society}, 55(4), 1959.

\bibitem{bossek2019evolving}
Jakob Bossek, Pascal Kerschke, Aneta Neumann, Markus Wagner, Frank Neumann, and Heike Trautmann.
\newblock Evolving diverse tsp instances by means of novel and creative mutation operators.
\newblock In {\em Proceedings of the 15th ACM/SIGEVO conference on foundations of genetic algorithms}, pages 58--71, 2019.

\bibitem{fruchterman1991graph}
Thomas~MJ Fruchterman and Edward~M Reingold.
\newblock Graph drawing by force-directed placement.
\newblock {\em Software: Practice and experience}, 21(11):1129--1164, 1991.

\bibitem{gent1996tsp}
Ian~P Gent and Toby Walsh.
\newblock The tsp phase transition.
\newblock {\em Artificial Intelligence}, 88(1-2):349--358, 1996.

\bibitem{gil2023evolving}
Francisco~J Gil-Gala, Marko Durasevi{\'c}, Mar{\'\i}a~R Sierra, and Ramiro Varela.
\newblock Evolving ensembles of heuristics for the travelling salesman problem.
\newblock {\em Natural Computing}, 22(4):671--684, 2023.

\bibitem{held1962dynamic}
Michael Held and Richard~M Karp.
\newblock A dynamic programming approach to sequencing problems.
\newblock {\em Journal of the Society for Industrial and Applied mathematics}, 10(1):196--210, 1962.

\bibitem{jin2023pointerformer}
Yan Jin, Yuandong Ding, Xuanhao Pan, Kun He, Li~Zhao, Tao Qin, Lei Song, and Jiang Bian.
\newblock Pointerformer: Deep reinforced multi-pointer transformer for the traveling salesman problem.
\newblock In {\em Proceedings of the AAAI Conference on Artificial Intelligence}, volume~37, pages 8132--8140, 2023.

\bibitem{bib2}
Lyle A.~McGeoch Johnson David~S. and Edward~E. Rothberg.
\newblock Asymptotic experimental analysis for the held-karp traveling salesman bound.
\newblock {\em roceedings of the 7th Annual ACM-SIAM Symposium on Discrete Algorithms}, 341, 1996.

\bibitem{joshi2022learning}
Chaitanya~K Joshi, Quentin Cappart, Louis-Martin Rousseau, and Thomas Laurent.
\newblock Learning the travelling salesperson problem requires rethinking generalization.
\newblock {\em Constraints}, 27(1):70--98, 2022.

\bibitem{kaplan2020scalinglawsneurallanguage}
Jared Kaplan, Sam McCandlish, Tom Henighan, Tom~B. Brown, Benjamin Chess, Rewon Child, Scott Gray, Alec Radford, Jeffrey Wu, and Dario Amodei.
\newblock Scaling laws for neural language models, 2020.

\bibitem{kool2018attention}
Wouter Kool, Herke Van~Hoof, and Max Welling.
\newblock Attention, learn to solve routing problems!
\newblock {\em arXiv preprint arXiv:1803.08475}, 2018.

\bibitem{kwon2020pomo}
Yeong-Dae Kwon, Jinho Choo, Byoungjip Kim, Iljoo Yoon, Youngjune Gwon, and Seungjai Min.
\newblock Pomo: Policy optimization with multiple optima for reinforcement learning.
\newblock {\em Advances in Neural Information Processing Systems}, 33:21188--21198, 2020.

\bibitem{larranaga1999genetic}
Pedro Larranaga, Cindy M.~H. Kuijpers, Roberto~H. Murga, Inaki Inza, and Sejla Dizdarevic.
\newblock Genetic algorithms for the travelling salesman problem: A review of representations and operators.
\newblock {\em Artificial intelligence review}, 13:129--170, 1999.

\bibitem{lecun2015deep}
Yann LeCun, Yoshua Bengio, and Geoffrey Hinton.
\newblock Deep learning.
\newblock {\em nature}, 521(7553):436--444, 2015.

\bibitem{lischka2024less}
Attila Lischka, Jiaming Wu, Rafael Basso, Morteza~Haghir Chehreghani, and Bal{\'a}zs Kulcs{\'a}r.
\newblock Less is more-on the importance of sparsification for transformers and graph neural networks for tsp.
\newblock {\em arXiv preprint arXiv:2403.17159}, 2024.

\bibitem{onwubolu2004optimizing}
Godfrey~C Onwubolu, BV~Babu, and Godfrey~C Onwubolu.
\newblock Optimizing cnc drilling machine operations: traveling salesman problem-differential evolution approach.
\newblock {\em New optimization techniques in engineering}, pages 537--565, 2004.

\bibitem{ralphs2003capacitated}
Ted~K Ralphs, Leonid Kopman, William~R Pulleyblank, and Leslie~E Trotter.
\newblock On the capacitated vehicle routing problem.
\newblock {\em Mathematical programming}, 94:343--359, 2003.

\bibitem{reinelt1995tsplib95}
Gerhard Reinelt.
\newblock Tsplib95.
\newblock {\em Interdisziplin{\"a}res Zentrum f{\"u}r Wissenschaftliches Rechnen (IWR), Heidelberg}, 338:1--16, 1995.

\bibitem{bib1}
Daniel Rosenkrantz, Richard Stearns, and Philip II.
\newblock An analysis of several heuristics for the traveling salesman problem.
\newblock {\em SIAM J. Comput.}, 6:563--581, 09 1977.

\bibitem{sanchis1990complexity}
Laura~A. Sanchis.
\newblock On the complexity of test case generation for np-hard problems.
\newblock {\em Information Processing Letters}, 36(3):135--140, 1990.

\bibitem{Jingyan:196322}
Jingyan Sui, Shizhe Ding, Xulin Huang, Yue Yu, Ruizhi Liu, Boyang Xia, Zhenxin Ding, Liming Xu, Haicang Zhang, Chungong Yu, and Dongbo Bu.
\newblock A survey on deep learning-based algorithms for the traveling salesman problem.
\newblock {\em Frontiers of Computer Science}, 19(6):196322, 2025.

\bibitem{vinyals2015pointer}
Oriol Vinyals, Meire Fortunato, and Navdeep Jaitly.
\newblock Pointer networks.
\newblock {\em Advances in neural information processing systems}, 28, 2015.

\bibitem{volgenant1982branch}
Ton Volgenant and Roy Jonker.
\newblock A branch and bound algorithm for the symmetric traveling salesman problem based on the 1-tree relaxation.
\newblock {\em European Journal of Operational Research}, 9(1):83--89, 1982.

\bibitem{wang2021game}
Chenguang Wang, Yaodong Yang, Oliver Slumbers, Congying Han, Tiande Guo, Haifeng Zhang, and Jun Wang.
\newblock A game-theoretic approach for improving generalization ability of tsp solvers.
\newblock {\em arXiv preprint arXiv:2110.15105}, 2021.

\bibitem{wang2024asp}
Chenguang Wang, Zhouliang Yu, Stephen McAleer, Tianshu Yu, and Yaodong Yang.
\newblock Asp: Learn a universal neural solver!
\newblock {\em IEEE Transactions on Pattern Analysis and Machine Intelligence}, 2024.

\bibitem{wu2023dataoptimizationdeeplearning}
Ou~Wu and Rujing Yao.
\newblock Data optimization in deep learning: A survey, 2023.

\bibitem{Bresson2021}
Bresson Xavier and Thomas Laurent.
\newblock The transformer network for the traveling salesman problem.
\newblock {\em arXiv preprint arXiv:2103.03012}, 2021.

\bibitem{yehuda2020s}
Gal Yehuda, Moshe Gabel, and Assaf Schuster.
\newblock It’s not what machines can learn, it’s what we cannot teach.
\newblock In {\em International conference on machine learning}, pages 10831--10841. PMLR, 2020.

\bibitem{yook2024cycleformer}
Jieun Yook, Junpyo Seo, Joon Huh, Han~Joon Byun, and Byung-ro Moon.
\newblock Cycleformer: Tsp solver based on language modeling.
\newblock {\em arXiv preprint arXiv:2405.20042}, 2024.

\bibitem{zhang2022learning}
Zeyang Zhang, Ziwei Zhang, Xin Wang, and Wenwu Zhu.
\newblock Learning to solve travelling salesman problem with hardness-adaptive curriculum.
\newblock In {\em Proceedings of the AAAI Conference on Artificial Intelligence}, volume~36, pages 9136--9144, 2022.

\end{thebibliography}

\medskip

%%%%%%%%%%%%%%%%%%%%%%%%%%%%%%%%%%%%%%%%%%%%%%%%%%%%%%%%%%%%
\newpage

\appendix
\title{Appendix}
\renewcommand{\thefigure}{S\arabic{figure}}  % 重新定义编号格式
\setcounter{figure}{0}
\renewcommand{\thetable}{S\arabic{table}}  % 重新定义编号格式
\setcounter{table}{0}
\section{Appendix}
\subsection{Proof of Theorem 1}

\textbf{Proof.}
According to the Remark~\ref{rem_1}, we know for the uniform distribution, the calculation of $\rho_n$ simplifies to a specific form
\[
\rho_n^* = \sum\limits_{c_i\in V}\frac{I(c_i)}{n}.
\]
Let \( r_k \) denote the distance from a given reference point to its \( k \)-th nearest-neighbor. For an instance with \( n \) nodes, a proportion \( \rho_n \) of the nodes are connected to their nearest-neighbors on the optimal hamiltonian tour. Therefore, the lower bound for the edges associated with these \( \rho_n \cdot n \) nodes is given by:
\begin{equation}
    \frac{1}{2}\rho_n \cdot n (r_1+r_2)+\theta_1,
\end{equation}
 while the lower bound for the edges associated with the other nodes is given by:
\begin{equation}
    \frac{1}{2}\rho_n \cdot n (r_2+r_3)+\theta_2,
\end{equation}
where $\theta_1$ and $\theta_2$ represent the deviation.

Therefore, we need to figure out the statistical distribution of the \( k \)-nearest-neighbor distances.

For a given reference point the absolute probability of finding its $k$ -th neighbour $(k<n)$ at a distance between $r_k$  and $r_k+d r_k$ from it is given by the probability that out of the $n-1$ random points (other than the reference point) distributed uniformly within the hypersphere of unit volume, exactly $k-1$ points lie within a concentric hypersphere of radius $r_k$ and at least one of the remaining $n-k$ points lie within the shell of internal radius $r_k$ and thickness $d r_k$. Therefore,
\begin{equation}
P(r_k)d r_k =\binom{n-1}{k-1}V_k^{k-1}\sum\limits_{q=1}^{n-k}\binom{n-k}{q}(1-V_k)^{n-k-q}(d V_k)^q,
\end{equation}
and $\mathbb{E}(r_k) = \int_0^R r_k P(r_k)$,
where $R$ is the radius of the $D$-dimensional hypersphere of unit volum.

Thus, we have
\begin{equation}
\begin{aligned}
    \mathbb{E}(r_k)=& \frac{\Gamma(\frac{D}{2}+1)^{\frac{1}{D}}}{\sqrt{\pi}}\binom{n-1}{k-1}(n-k)\int_0^1V_k^{k+\frac{1}{D}-1}(1-V_k)^{n-k-1} dV_k\\
    =& \frac{\Gamma(\frac{D}{2}+1)^{\frac{1}{D}}}{\sqrt{\pi}}\binom{n-1}{k-1}(n-k) B(k+\frac{1}{D},n-k)\\
    =& \frac{\Gamma(\frac{D}{2}+1)^{\frac{1}{D}}}{\sqrt{\pi}}\frac{\Gamma(k+\frac{1}{D})}{\Gamma(k)}\frac{\Gamma(n)}{\Gamma(n+\frac{1}{D})},
\end{aligned}
\end{equation}
and
\begin{equation}
\begin{aligned}
    \mathbb{E}(r_k^2) &= \frac{\Gamma(\frac{D}{2}+1)^{\frac{2}{D}}}{\pi}\binom{n-1}{k-1}(n-k)B(k+\frac{2}{D},n-k)\\
    & = \frac{\Gamma(\frac{D}{2}+1)^{\frac{2}{D}}}{\pi}\frac{\Gamma(k+\frac{2}{D})}{\Gamma(k)}\frac{\Gamma(n)}{\Gamma(n+\frac{2}{D})}.\\
\end{aligned}
\end{equation}

When taking $D=2$ and using Stirling’s approximation, we get $\mathbb{E}(r_k)=\frac{1}{\sqrt{n\pi}}\frac{\Gamma(k+\frac{1}{D})}{\Gamma(k)}$ and $\mathbb{E}(r_k^2)=\frac{1}{n\pi}\frac{\Gamma(k+\frac{2}{D})}{\Gamma(k)}$. Let $r_{k,1}, \cdots, r_{k,n}$ be $n$ identically distributed copies of the random variable $r_k$,
thus
\[
\mathbb{E}[\vert\sqrt{\frac{1}{n} }\sum\limits_{i=1}^n r_{k,i}-\sqrt{n}E(r_k)\vert]=O(\frac{1}{n}).
\]
Given that $\lim\limits_{n\rightarrow \infty} \frac{l_{\text{opt}}}{\sqrt{n}}=\beta$ and  $\lim\limits_{n\rightarrow\infty}\frac{1}{2}\sqrt{\frac{1}{n}}\cdot[\rho_n\sum\limits_{i=1}^n (r_{1,i}+r_{2,i})+(1-\rho_n)\sum\limits_{i=1}^n(r_{2,i}+r_{3,i})]\leq \beta $, a.s.,

we have 
\[
\lim\limits_{n\rightarrow\infty}\frac{5}{8}\rho_n +\frac{27}{32}(1-\rho_n)\leq\beta,~\text{a.s.}
\]
and we then concluded that $\rho_n\geq\frac{27-32\beta}{7}~\text{a.s.}$, as $n\rightarrow \infty$. 

\qed
\subsection{Proof of Theorem 3}
\textbf{Proof.} Define the language $L$ as follows: Given an instance $I$ of the two-dimensional Euclidean TSP and a value $\delta \in (0,1)$, if $\rho(I)\leq \delta$, then $I\in L$. Otherwise $I\in L^c$. $L$ is NP-hard because it depends on solving the TSP itself. And the classification task $T(L)$ introduced by $L$ is NP-hard, i.e., given an instance $I$, decide if $I\in L$.

Assume that there exists an efficient complete sampler $S_L$ for $L$. Since $L$ is complete, the classification task $T(S_L)$ induced by $S_L$, i.e., given an instance $I$ generated by $S_L$, determine if $I\in L$, is equal to the task $T(L)$ introduced by $L$. Given a $I$ generated by $S_L$, and $I\in L$, let $sq$ be the sequence of random bits used by $S_L$ to generate $I$. Since $S_L$ runs in polynomial time, it must use at most polynomial number of random bits. Then $\forall I\in S_L$ , there exists a string $sq$ with length polynomial in $\vert I\vert$. Therefore a deterministic Turing machine that given $I$ and $sq$ can verify in polynomial time if $I\in L$. So $T(S_L)\in NP$. Similarly, we can show that $T(S_L) \in coNP$. Thus $T(S_L)\in NP\bigcap coNP$. Then, we have $T(L)$ in $NP\bigcap coNP$. However, $T(L)\in NP-hard$, and if $T(L)\in NP\bigcap coNP$, then $NP=coNP$. This leads to a contradiction.
\qed

\subsection{Proof of Theorem 4}

\textbf{Proof.} Using proof by contradiction, assume that the efficient algorithm covering exists. We can simply obtain a polynomial-time exact solving algorithm by sequentially running $\mathcal{A}_i$. Then $\mathcal{P}\in P$. Since \( P \in NPC \), this would imply \( P = NP \), which is a contradiction. 
\qed

\subsection{Proof of Theorem 5}
\textbf{Proof.} The proof idea follows the proof of Theorem 4.
\qed

\subsection{Introduction to Barabási-Albert Model}
\label{Appendix-BA}
The Barabási-Albert model is a fundamental framework for generating scale-free networks characterized by growth and preferential attachment. This model, also referred to as the scale-free model, operates as follows:

Initially, the network consists of $m_0$ nodes, connected arbitrarily, ensuring each node has at least one link. The network evolves through two primary mechanisms:

\begin{itemize}
\item \textbf{Growth.} At each time step, a new node is introduced to the network with $m$ (where \( m \leq m_0 \)) links. These links connect the new node to \( m \) existing nodes in the network.

\item \textbf{Preferential Attachment.} The probability \( P(k_i) \) that a new node will connect to an existing node \( i \) is proportional to the degree \( k_i \) of node \( i \). This probability is given by:
\[
P(k_i) = \frac{k_i}{\sum_{j} k_j}. 
\]
\end{itemize}
Preferential attachment is a probabilistic process, allowing new nodes to connect to any existing node, regardless of whether it is a highly connected hub or a node with few links. 

This model effectively captures the dynamics of real-world networks, where new nodes tend to connect to already well-connected nodes, leading to a scale-free topology.

\subsection{Experimental Details}
\label{appendix-ed}
We use the model by Bresson et al.~\cite{Bresson2021} as the base solver. It is a Transformer-based architecture. The transformer is learned by reinforcement learning. Hence, no TSP solutions/approximations are required. The solver has quadratic complexity $O(n^2L)$, where $L$ is the depth of the models. Details of the network architecture and training process are provided in \textbf{Table~\ref{table-4}}. 

The base solvers involved in the comparison all have open-source model parameters. The decoding type of beam search is with the beam width uniformly set to 1000. During fine-tuning, we replaced one-fourth of the original training set (uniformly distributed) with sample instances constructed using the drilling model examples. The number of fine-tuning steps for different solvers is provided in \textbf{Table~\ref{table-4}}. It should be noted that for POMO~\cite{kwon2020pomo} and Pointerformer~\cite{jin2023pointerformer}, the authors adopted an optional unit square transformations scheme during the inference phase. We omitted this step during comparisons to ensure fairness with other models.

The RUE test set contains 1280 instances with the node size of 50, derived from the test set used in prior works~\cite{vinyals2015pointer,yook2024cycleformer}. However, we have revised the exact solutions for this test set. For the other two types of instances with a node size of 50, we have used our own generated datasets, each consisting of 1000 instances.

\begin{table}[h]
  \caption{Details of the neural network architecture and training process}
  \centering
  \scriptsize
  % \hspace*{-1.3cm}
  \resizebox{0.9\textwidth}{!}{\begin{tabular}{lccccc}
    \toprule
 Base Solver  & Batch Size & Total TSP samples for training (fine-tuning)  \\
 \midrule
TSP-GNN & 512 & $1.28\times 10^7$ \textcolor{pnasbluetext}{( $6.40\times 10^6$)} \\
TSP-Transformer & 512 & $2.08\times 10^9$ \textcolor{pnasbluetext}{$(1.28\times 10^7)$} \\
POMO & 128 & $2.00\times 10^8$ \textcolor{pnasbluetext}{$(1.28\times 10^7)$} \\
Pointerformer & 64 & $3.00\times 10^8$ \textcolor{pnasbluetext}{$(1.28\times 10^7)$} \\
Cycleformer & 80 & $1.50\times 10^8$ $\textcolor{pnasbluetext}{(4.00\times 10^6)}$\\
    \bottomrule
  \end{tabular}}
\label{table-4}
\end{table}

\subsection{Convolutional distribution Instances}
Wang \emph{et al.} proposed a new method called ASP~\cite{wang2024asp}: Adaptive Staircase Policy Space Responce Oracle (PSRO), which aims to address the generalization issues faced by previous neural solvers and provide a "universal neural solver" that can be applied to a wide range of problem distributions and scales. They generate data by randomly sampling $x\in\mathbb{R}^2$ from the unit square, and sampling $y\in\mathbb{R}^2$ from $N(0,\Sigma)$ where $\Sigma\in\mathbb{R}^{2\times 2}$ is a diagonal matrix whose elements are sampled from $[0,\lambda]$ and $\lambda\sim U(0,1)$. Next, a two-dimensional coordinate is generated by $z=x+y$. This describes a convolutional distribution which is the convolution of the uniform distribution and the Gaussian distribution.

\begin{figure}[H]
    \centering
    \includegraphics[width=0.95\linewidth]{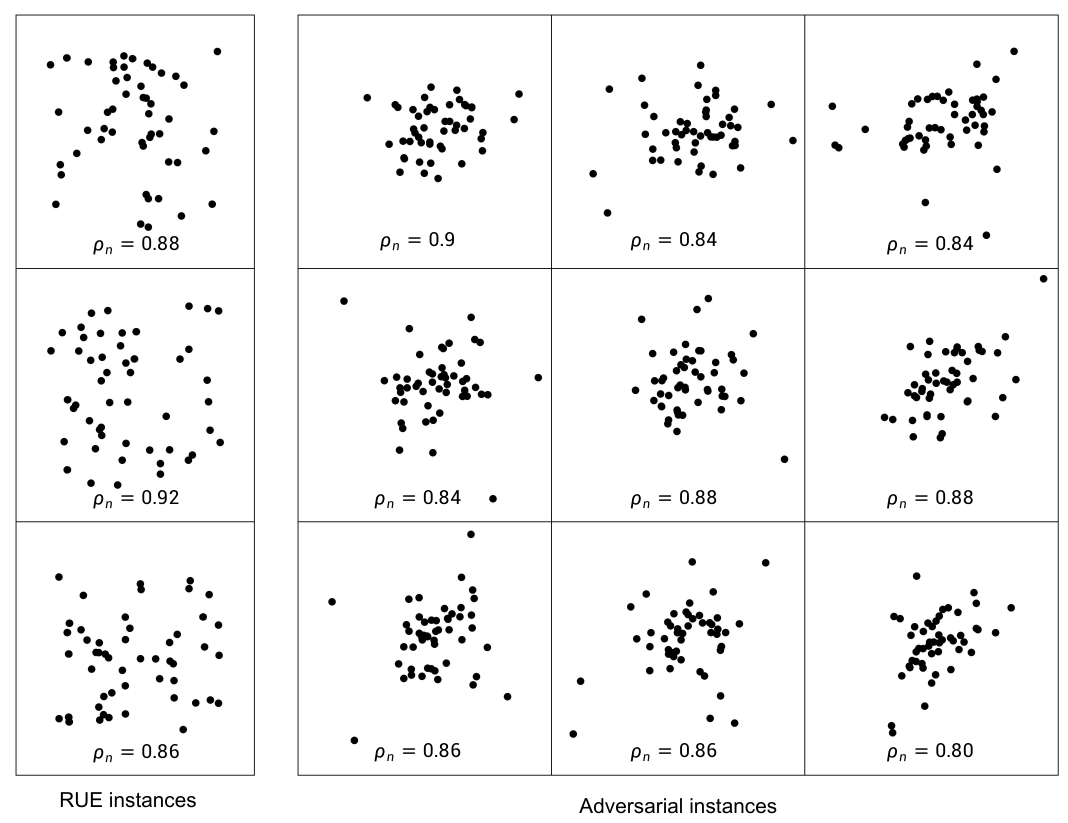}
    \caption{Visualization of some adversarial instances}
    \label{fig:Fig6}  

\end{figure}

We analyzed the nearest-neighbor density of these instances and found that they did not exhibit significantly low-density characteristics (\textbf{Table~\ref{table-5}}). Interestingly, however, compared to a uniform distribution, there was indeed a noticeable decrease. This further demonstrates the effectiveness of our metric in evaluating the difficulty of instances for neural solvers. Additionally, we visualized some of the instances (\textbf{Fig.~\ref{fig:Fig6}}), and interestingly, their distributions revealed regions of both sparsity and density. This closely resembles the examples we constructed based on scale-free complex networks.

\begin{table}[h]
  \caption{The estimation of nearest density $\rho_n$ for adversarial instances }
  \centering
  \scriptsize
  % \hspace*{-1.3cm}
  \resizebox{0.45\textwidth}{!}{\begin{tabular}{lccc}
    \toprule

 Node Size & \multicolumn{3}{c}{The Nearest Density $\rho_n$}  \\
      & Adversarial Instances \\
    \midrule
    20 & 0.8506 \\
    30 & 0.8602 \\
    40 & 0.8619 \\
    50 & 0.8642 \\
    \bottomrule
  \end{tabular}}
\label{table-5}
\end{table}

\end{document}